\documentclass[10pt,letterpaper]{article}
\usepackage{natbib}
\usepackage{aaai22}
\usepackage{times}
\usepackage{helvet}
\usepackage{courier}
\usepackage{tabularx}
\usepackage{xparse}
\usepackage{amsmath}
\usepackage{amsthm}
\usepackage{amssymb}
\usepackage{amsfonts}
\usepackage{xspace}
\usepackage{relsize}
\usepackage{graphicx}
\usepackage{multirow}
\usepackage{url}
\usepackage{comment}
\usepackage{math_commands}
\usepackage[svgnames]{xcolor}

\nocopyright

\usepackage{booktabs}   

\usepackage{xr}

\usepackage{stackengine}
\stackMath
\newcommand\tsup[2][2]{%
 \def\useanchorwidth{T}%
  \ifnum#1>1%
    \stackon[-.5pt]{\tsup[\numexpr#1-1\relax]{#2}}{\scriptscriptstyle\sim}%
  \else%
    \stackon[.5pt]{#2}{\scriptscriptstyle\sim}%
  \fi%
}

\renewcommand{\tilde}[1]{\tsup[1]{#1}}
\newcommand{\uline}[1]{\underline{#1}}

\newcommand{\function}[1]{\textsc{#1}}

\newcommand{\mycolor}[2]{\textcolor{#1}{#2}}

\newcommand{\red}[1]{\mycolor{red}{#1}}

\def\_{\\[-0.3em]}

\newtheorem{defi}{Definition}
\newtheorem{ex}{Example}
\newtheorem{theo}{Theorem}

\newtheorem{conv}{Convention}

\makeatletter
\let\@myref\ref

\newcommand{\refsec}[1]{Sec.\,\@myref{#1}}
\newcommand{\refseq}[1]{Sec.\,\@myref{#1}}
\newcommand{\refig}[1]{Fig.\,\@myref{#1}}
\newcommand{\reftbl}[1]{Table \@myref{#1}}
\newcommand{\refstep}[1]{Step \@myref{#1}}
\newcommand{\refalgo}[1]{Alg.\,\@myref{#1}}
\newcommand{\refchap}[1]{Chap.\,\@myref{#1}}
\newcommand{\reflst}[1]{List \@myref{#1}}
\newcommand{\refeq}[1]{Eq.\,\@myref{#1}}

\newcommand{\reftheo}[1]{Thm.\,\@myref{#1}}
\newcommand{\refline}[1]{line\,\@myref{#1}}
\newcommand{\refdef}[1]{Def.\, \@myref{#1}}
\newcommand{\refex}[1]{Example\,\@myref{#1}}
\newcommand{\refconv}[1]{Conv.\,\@myref{#1}}
\newcommand{\reffact}[1]{Fact.\,\@myref{#1}}

\newcommand{\refsecs}[2]{Sec.\,\@myref{#1}-\@myref{#2}}
\newcommand{\refseqs}[2]{Sec.\,\@myref{#1}-\@myref{#2}}
\newcommand{\refigs}[2]{Fig.\,\@myref{#1}-\@myref{#2}}
\newcommand{\reftbls}[2]{Tables \@myref{#1}-\@myref{#2}}
\newcommand{\refsteps}[2]{Steps \@myref{#1}-\@myref{#2}}
\newcommand{\refalgos}[2]{Alg.\,\@myref{#1}-\@myref{#2}}
\newcommand{\refchaps}[2]{Chap.\,\@myref{#1}-\@myref{#2}}
\newcommand{\reflsts}[2]{Lists \@myref{#1}-\@myref{#2}}
\newcommand{\refeqs}[2]{Eq.\,\@myref{#1}-\@myref{#2}}
\newcommand{\refpages}[2]{p.\pageref{#1}-\@myref{#2}}
\newcommand{\reftheos}[2]{Thm.\,\@myref{#1}-\@myref{#2}}
\newcommand{\reflines}[2]{line\,\@myref{#1}-\@myref{#2}}
\newcommand{\refdefs}[2]{Def.\,\@myref{#1}-\@myref{#2}}
\newcommand{\refexs}[2]{Example\,\@myref{#1}-\@myref{#2}}
\newcommand{\refconvs}[2]{Conv.\,\@myref{#1}-\@myref{#2}}
\newcommand{\reffacts}[2]{Facts.\,\@myref{#1}-\@myref{#2}}

\makeatother

\newcounter{list}[section]

\makeatletter
\@ifundefined{citet}{
  \NewDocumentCommand{\citet}{o m}{%
    \IfNoValueTF{#1}%
      {\citeauthor{#2} (\citeyear{#2})}
      {\citeauthor{#2} (\citeyear[#1]{#2})}%
  }
}{}
\@ifundefined{citep}{
  \NewDocumentCommand{\citep}{o m}{%
    \IfNoValueTF{#1}%
      {\cite{#2}}
      {\cite[#1]{#2}}%
  }
}{}
\makeatother

\newcommand{\todo}[1]{\red{\textbf{#1}}}

\newlength{\maxwidth}
\newcommand{\algalign}[2]
{\makebox[\maxwidth][r]{$#1{}$}${}#2$}

\makeatletter

\newcommand{\newheuristic}[2]{%
 \def#1{%
  \ifmmode%
  h^\text{#2}\xspace%
  \else%
  \text{#2}\xspace%
  \fi%
 }%
}

\newheuristic{\lmcut}{LMcut}
\newheuristic{\mands}{M\&S}
\newheuristic{\pdb}{PDB}
\newheuristic{\ff}{FF}
\newheuristic{\ce}{CEA}
\newheuristic{\cg}{CG}
\newheuristic{\ad}{add}
\newheuristic{\hmax}{max}
\newheuristic{\lc}{LC}
\newheuristic{\blind}{blind}

\newcommand{\newlearnedheuristic}[2]{%
 \def#1{%
  \ifmmode%
  H^\text{#2}\xspace%
  \else%
  \text{#2}\xspace%
  \fi%
 }%
}

\newlearnedheuristic{\Hlmcut}{LMcut}
\newlearnedheuristic{\Hmands}{M\&S}
\newlearnedheuristic{\Hpdb}{PDB}
\newlearnedheuristic{\Hff}{FF}
\newlearnedheuristic{\Hce}{CEA}
\newlearnedheuristic{\Hcg}{CG}
\newlearnedheuristic{\Had}{add}
\newlearnedheuristic{\Hmax}{max}
\newlearnedheuristic{\Hlc}{LC}
\newlearnedheuristic{\Hblind}{blind}

\newcommand{\newUnitCostHeuristic}[2]{%
 \def#1{%
  \ifmmode%
  \hat{h}^\text{#2}\xspace%
  \else%
  \text{#2}\xspace%
  \fi%
 }%
}

\newUnitCostHeuristic{\lmcuto}{LMcut}
\newUnitCostHeuristic{\mandso}{M\&S}
\newUnitCostHeuristic{\ffo}{FF}
\newUnitCostHeuristic{\ceo}{CEA}
\newUnitCostHeuristic{\cgo}{CG}
\newUnitCostHeuristic{\ado}{add}
\newUnitCostHeuristic{\gco}{GoalCount}
\newUnitCostHeuristic{\lco}{LC}

\makeatother

\makeatletter
\renewcommand{\ref}[1]{\@myref{#1}\todo{(Do not use ``ref'' directly!)}}
\makeatother

\allowdisplaybreaks

\title{Analytical Conjugate Priors for Subclasses of Generalized Pareto Distributions}

\author{Masataro Asai}
\affiliations {MIT-IBM Watson AI Lab, masataro.asai@ibm.com}

\begin{document}
\maketitle
\begin{abstract}
This article is written for pedagogical purposes aiming at practitioners
trying to estimate the \emph{finite support of continuous probability distributions},
i.e., the minimum and the maximum of a distribution defined on a finite domain.
Generalized Pareto distribution $\gp(\theta,\sigma,\xi)$ is a three-parameter distribution which plays a key role in
Peaks-Over-Threshold framework for tail estimation in Extreme Value Theory.
Estimators for GP often lack analytical solutions and
the best known Bayesian methods for GP involves numerical methods.
Moreover, existing literature focuses on estimating the scale $\sigma$ and the shape $\xi$,
lacking discussion of the estimation of the location $\theta$ which is the lower support of (minimum value possible in) a $\gp$.
To fill the gap, we analyze four two-parameter subclasses of $\gp$
whose conjugate priors can be obtained analytically,
although some of the results are known.
Namely, we prove the conjugacy for
$\xi>0$ (Pareto),
$\xi=0$ (Shifted Exponential),
$\xi<0$ (Power),
and $\xi=-1$ (Two-parameter Uniform).
\end{abstract}

\section{Introduction}

Regular statistics are typically built around the Central Limit Theorem,
which deals with the limit behavior of a sum/average of multiple samples.
In contrast, a branch of statistics called \emph{Extreme Value Theory} \cite[EVT]{beirlant2004statistics,dehaan2006extreme}
is built around the \emph{Extremal Limit Theorem},
a theorem that describes the limit behavior of the maximum of multiple samples.
EVTs has been historically used for predicting the behavior of safety-critical natural phenomena
whose best / worst case behaviors matter.
For example, in hydrology, the annual maximum discharge (water level) of a river affects
the height of the embankment that is necessary for ensuring the safety.

In recent years,
the AI community has been increasing its focus on
the safety of machine learning system as well as
hybrid neuro-symbolic systems that
leverage the deterministic guarantees of logical correctness / optimality in the symbolic system
that is missing in the purely connectionist approaches.
Theoretically guaranteed safety of predictions produced by a machine learning system is paramount
in safety-critical applications such as autonomous driving.
A safety criteria is typically provided in a form of upper / lower bounds and
all symbolic constraint optimization algorithms, including Mixed Integer Linear Programming,
MAX-SAT solvers \cite{davies2013solving}, or Automated Planning \cite{pddlbook},
maximize or minimize an objective function guided usually by a lower / upper bound of a solution.

Averages are useful, but extrema deserve more attention.
While the mainstream ML focuses on the \emph{most likely} behavior (e.g., a MAP estimate) based on CLT,
real-world safety-critical applications must know the model's highly \emph{unlikely} limit behaviors that CLT does not capture.
For example,
we are not only interested in the average travel time to the office,
but also in its worst case (not to be late for a meeting)
and its best case (to know how good my route is; to take the risk to improve the plan further).
It even makes sense in creative applications like text-to-image models \cite[DALL-E]{dalle,dalle2}:
A novel art emerges from an exaggeration toward the extremes, not from regression to the incompetent norms.

To model these behaviors,
one must understand the \emph{least likely} best/worst case behavior
that can be found at the edges of a distribution, which are formally called \emph{tail distributions}.
In decision making tasks that are traditionally handled by symbolic AI,
these rare, least likely values are often precisely what we want to know.
Unfortunately, existing machine learning theories are useless in predicting such an unlikely behavior
due to the ubiquitous reliance on CLT, which only models the average / \emph{most likely} behaviors.
Modeling the limit behaviors requires Extreme Value Theory that provides a strong statistical justification for
analyzing and predicting the tail distributions.

This article is written for readers not familiar with EVTs and are interested in the prediction of limit behaviors.
To appeal to these audience,
we do not provide a lengthy survey on the entire field of EVTs,
but rather provide several known results whose proofs are difficult to find but are easy to apply to existing tasks.
More concretely,
we provide several analytically available conjugate prior distributions for
a limited subset of Generalized Pareto distribution,
a distribution that derives from Extreme Value Theorem Type 2.

\section{Preliminary}

\begin{defi}
 A \emph{probability distribution} of a random variable (RV) $\rx$ defined on a set $X$
 is a function $f$ from a value $x\in X$ to $f(x)\in \R^{0+}$
 which satisfies $1=\sum_{x\in X} f(x)$.
 $f(x)=p(\rx=x)=p(x)$ is called a probability mass/density of an event $\rx=x$.
 $f$ is called a probability mass/density function (PMF/PDF).
 When $F(x)=p(\rx\leq x)$, $F$ is called a cumulative distribution function (CDF).
\end{defi}

\begin{defi}
 A \emph{support} of a function $f$ is a set where it is non-zero, $\function{supp}(f)=\braces{x| f(x)\not=0}$.
 It often assumes non-negative functions including probability distributions.
\end{defi}

\begin{defi}
 \label{defi:joint}
 A \emph{joint distribution} $p(\rx, \ry)$ is a function $X\times Y\to \R^+$
 satisfying $1=\sum_{(x,y)\in X\times Y} p(x,y)$,
 $p(x)=\sum_{y\in Y} p(x,y)$, and $p(y)=\sum_{x\in X} p(x,y)$,
 given $p(\rx)$, $p(\ry)$.
\end{defi}

\begin{defi}
 $f(x,y)=p(\rx=x,\ry=y)=p(x,y)$ is called a probability mass/density of observing $\rx=x$ and $\ry=y$ at the same time
 (also written as $p(\rx=x\land\ry=y)$).
\end{defi}

\begin{defi}
 \label{defi:conditional}
 A \emph{conditional distribution}
 $p(\rx\mid \ry)$ is $\frac{p(\rx,\ry)}{p(\ry)}$.
\end{defi}

\begin{defi}
 RVs $\rx, \ry$ are independent when $p(\rx,\ry)=p(\rx)p(\ry)$,
 denoted by $\rx \perp \ry$.
\end{defi}

\begin{defi}
 RVs $\rx, \ry$ are independent and identically distributed (i.i.d) when
 $p(\rx)=p(\ry)$ and $\rx\perp \ry$.
\end{defi}

\begin{defi}
 \label{def:expectation}
 An \emph{expectation} of a quantity $g(x)$ over $p(\rx)$
 is defined as $\E_{x\sim p(\rx)}g(x)=\E_{p(x)}g(x)=\sum_{x\in X}p(x)g(x)$
 if $\sum_{x\in X}p(x)|g(x)|<\infty$. It does not exist otherwise.
\end{defi}

\begin{defi}
 Functions $f_n$ \emph{converges pointwise} to $f$ ($f_n\to[n\to\infty] f$)
 when
 $\forall x; \forall \epsilon; \exists n; |f_n(x)- f(x)|<\epsilon$.
\end{defi}

\begin{defi}
 RVs $\rx_n$ converges to a RV $\rx$ \emph{in distribution}
 if $p\parens{\rx_n}=f_n \to[n\to\infty] f=p\parens{\rx}$,
 denoted as $\rx_n\to[D] \rx$.
\end{defi}

Central Limit Theorem \cite[CLT]{laplace1812centrallimittheorem} famously states that
the average $\ry_n$ of \iid RVs $\rx_1,\ldots \rx_n\sim p(\rx)$
with a finite expectation $\mu=\E[\rx]$ and a finite variance $\sigma^2=\E[(\rx-\mu)^2]$
``asymptotically follows'', i.e.,
converges in distribution to a Gaussian distribution.

\begin{defi}[Gaussian distribution]
\begin{align*}
 \N(\rx\mid\mu,\sigma^2)&=\frac{1}{\sqrt{2\pi\sigma^2}}\exp{-\frac{(\rx-\mu)^2}{2\sigma^2}}.
\end{align*}
\end{defi}

\begin{theo}[CLT]
 $\ry_n=\frac{1}{n}\sum_i (\rx_i-\mu) \to[D] \ry\sim\N(0,\sigma^2)$. \label{theo:clt}
\end{theo}

\begin{defi}
 A Dirac's delta distribution $\delta(\mu)$ is a pointwise limit
 $\N(\rx\mid\mu,\sigma^2)\to[\sigma^2\downto 0]\delta(\rx\mid\mu)$ which satisfies
\begin{align*}
 \delta(\rx\mid\mu)&=
 \left\{
 \begin{array}{cl}
  \infty& \rx=\mu,\\
  0 & \text{otherwise}.
 \end{array}
 \right.
\end{align*}
\end{defi}

Extreme Value Theorem Type 1 /
Fisher--Tippett--Gnedenko theorem /
Extremal Limit Theorem \cite{fisher1928limiting,gnedenko1943limiting} states that,
given \iid RVs normalized by an appropriate series $a_n>0$ and $b_n$,
their maximum converges in distribution to a \emph{Extreme Value Distribution} (EVD).
This theorem is often used for modeling block (periodic) maxima of time-series data,
such as in hydrology \cite{beirlant2004statistics,dehaan2006extreme}.

\begin{defi}[Extreme Value Distribution]
 \label{def:evd}
 \begin{align*}
  \EVD(\rx\mid\mu,\sigma,\gamma)
  &=
  \left\{
  \begin{array}{ll}
   \exp \parens{ -(1+\gamma \frac{\rx-\mu}{\sigma})^{-\frac{1}{\gamma}}} & (\gamma\not= 0)\\
   \exp \parens{ -e^{-\frac{\rx-\mu}{\sigma}}}                           & (\gamma\not= 0)
  \end{array}
  \right.
 \end{align*}
\end{defi}

\begin{theo}[Fisher--Tippett--Gnedenko]
 Let $\rx_1,\ldots \rx_n \sim p(\rx)$ be i.i.d. RVs defined on $X$.
 When $\ry_n=\frac{\max_i \rx_i - a_n}{b_n} \to[D] \ry$
 for some $a_n>0$ and $b_n$ and $p(\ry)$ is not degenerate (e.g., $\delta$),
 then
 $\ry\sim \EVD(0,1,\gamma)$ for some $\gamma$.
\end{theo}

Extreme Value Theory Type 2 /
Pickands--Balkema--de~Haan theorem \cite{pickands1975statistical,balkema1974residual} states that
\iid RVs exceeding a sufficiently high threshold $\theta$
asymptotically follows a Generalized Pareto (GP) distribution.
This theorem is used for the Peaks-Over-Threshold analysis.

\begin{defi}[Generalized Pareto Distribution]
 \label{def:gp}
 \begin{align*}
  \gp(\rx\mid\theta,\sigma,\xi)
  &=
  \left\{
  \begin{array}{ll}
   \frac{1}{\sigma}\parens{1+\xi\frac{\rx-\theta}{\sigma}}^{-\frac{\xi+1}{\xi}}& (\xi\not=0)\\
   \frac{1}{\sigma}\exp \parens{-\frac{\rx-\theta}{\sigma}}          & (\xi=0)
  \end{array}
  \right.
 \end{align*}
\end{defi}

\begin{figure}[tb]
 \includegraphics[width=\linewidth]{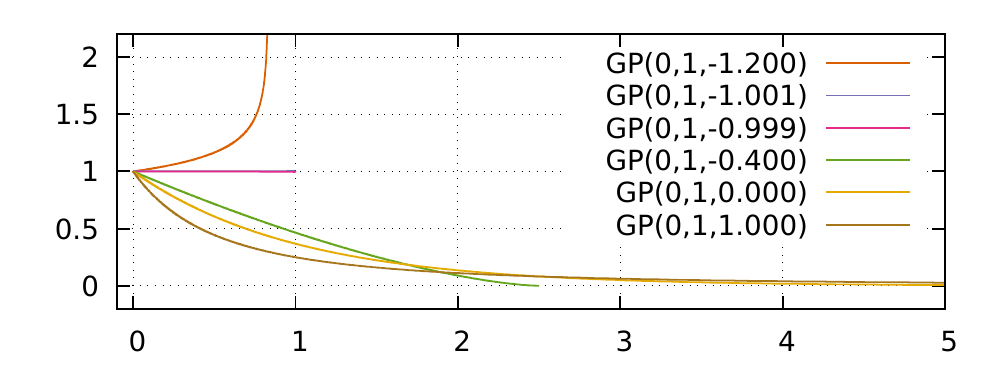}
 \caption{Generalized pareto distribution $\gp(0,1,\xi)$.}
 \label{fig:gp}
\end{figure}

\begin{theo}[Pickands--Balkema--de~Haan]
 Let $\rx_1,\ldots \rx_n \sim p(\rx)$ be i.i.d. RVs defined on $X$
 and $\rx_{k,n}=\theta$ be their top-$k$ element.
 As $n\to \infty$, $k\to \infty$, $\frac{k}{n}\to 0$ ($k\ll n$), then
 $p(\rx\mid\rx>\theta) \to \gp(\rx\mid \theta,\sigma,\xi)$
 for some $\sigma$ and $\xi$.
\end{theo}

$\theta$, $\sigma$ and $\xi$ are called the location, the scale and the shape parameter.
It has a support $\rx\in [\theta,\theta-\frac{\sigma}{\xi}]$ when $\xi<0$, otherwise $\rx\in [\theta,\infty]$.
The shape dictates the tail behavior:
$\xi>0$ corresponds to a heavy-tailed distribution,
$\xi<0$ corresponds to a short-tailed distribution,
and $\xi=0$ corresponds to a shifted Exponential distribution.
Pareto, Exponential, Power, and Uniform distributions
(\reftbl{tbl:gp-subclasses}) are special cases of GP distribution.
They share the characteristics that the density is 0 below a threshold $\theta$.

\begin{table}[tb]
 \centering
 \begin{tabular}{|l|c|c|c|}\toprule
  Special cases                    & $\theta$     & $\sigma$        & $\xi$             \\\midrule
  $\rx\sim \pareto(\alpha,l)$      & $l>0$        & ${l}/{\alpha}$  & ${1}/{\alpha}>0$  \\
  $\rx\sim \lomax(\alpha,l)$       & $0$          & ${l}/{\alpha}$  & ${1}/{\alpha}>0$  \\
  $\rx\sim \mathrm{Exp}(\alpha,l)$ & $l$          & ${1}/{\alpha}$  & $0$               \\
  $-\rx\sim \power(a,b)$           & $-a<0$       & ${a}/{b}$       & $-{1}/{b}<0$      \\
  $1/\rx\sim \ipareto(\alpha,l)$   & $-{1}/{l}<0$ & ${1}/{l\alpha}$ & $-{1}/{\alpha}<0$ \\
  $\rx\sim U(l,u)$                 & $l$          & $u-l$           & $-1$              \\
  \bottomrule
 \end{tabular}
 \caption{Special cases of Generalized Pareto distribution.}
 \label{tbl:gp-subclasses}
\end{table}

A typical application of GP is as follows:
Given a set of time-series data $(t_i,x_i)$,
extract a subset whose $x_i$ exceeds a certain sufficiently high threshold $\theta$, such as the top 5\% element.
Then $x_i$ in the subset, as well as the future exceeding data, follow a GP distribution.
GP is an accurate approximation when
we ignore almost all data ($\frac{k}{n}\to 0$)
while retaining enough data ($k\to \infty$)
as $n\to \infty$ and $\theta\to \sup X$.

Estimating the parameters of $\gp$ is known to be difficult.
First of all, the maximum likelihood estimator (MLE)
for $\sigma$ and $\xi$ of $\gp(0,\sigma,\xi)$ \cite{smith1987estimating} has many issues.
MLE of GP does not have a closed form solution except for the solution of $\sigma$ limited to $\xi=0$.
When $\xi<-1$, it does not exist because the log likelihood can reach $\infty$.
It is asymptotically normal and efficient only for $\xi>-0.5$.
Solving it requires a numerical method, such as the Newton-Raphson iteration.
\citet{hosking1987parameter} showed that
the iteration sometimes does not converge when $k$ is small or $\xi<0$, and
even if it converges, the solution can be inaccurate if $k$ is small.

Due to these issues, several consistent estimators
(i.e., an estimator that converges in probability to the true value as $k\to\infty$)
have been developed but still with varying degrees of limitations.
Hill's estimator \cite{hill1975simple} applies a limiting assumption to the maximum likelihood estimator
in order to obtain a closed form.
Smoothed Hill's estimator \cite{resnick1997smoothing} reduces its variance.
\citet{fraga2001location} addresses its location sensitivity.
Method of Moments (MOM) and Probability-Weighted Moments (PWM) \cite{hosking1987parameter}
do not have the limiting assumption, have closed forms, and are more accurate,
but they require $\xi<0.5$ to exist, and $\xi>0$ to be consistent,
because they rely on the existence of the zeroth and the first moment (i.e., the mean and the variance).
The Elemental Percentile Method (EPM) \cite{castillo1995method,castillo1997fitting}
addresses the restriction on $\xi$ in the existing methods, but it also involves a numerical method (Newton-Raphson),
and is better than MOM/PWM only when $\xi<0$.

Several Bayesian estimators are also available.
\citet{diebolt2005quasi} proposed a Bayesian quasi-conjugate prior,
but this is limited to $\xi>0$ and also requires a numerical method (Gibbs sampling).
\citet{sharpe2021estimation} proposed a method based on so-called
Bayesian Reference Intrinsic (BRI) approach \cite{bernardo2002bayesian}.
\citet{vilar2007pareto} proposed Generalized Inverse Gaussian distribution as a conjugate prior for Pareto distribution,
again limited to $\xi>0$.
\citet{sharpe2021estimation} also reproduced \citet{vilar2007pareto}
with a simplified independence assumption and Jeffery's improper prior.

\section{Bayesian Reasoning}

While existing literature focuses on the estimation of the tail index $\xi$
and the scale $\sigma$, not much focus is spent on the estimation of $\theta$.
However, estimating $\theta$ is important for predicting the minimum/maximum value that a random variable can take.
We are especially interested in the conjugate priors for the lower bound / threshold parameter $\theta$,
as well as the upper bound $\mu-\frac{\sigma}{\xi}$ that exists when $\xi<0$ (short-tailed distribution).
We also target the audience that are interested in a baseline formulation
that is easy to implement, is good enough for practical applications, or when absolute accuracy can be sacrificed.
We thus focus on deriving conjugate prior distributions for the special cases listed in \reftbl{tbl:gp-subclasses}.

We begin by the basic terminologies.

\begin{defi}
 Distributions $p(\rx_1)=f(\rx_1, \theta_1,\ldots,\theta_N)$ and $p(\rx_2)=f(\rx_2, \phi_1,\ldots,\phi_N)$ are of the \emph{same family} if
 they have the same functional form $f$ only differing in the parameters $\theta_1,\ldots,\theta_N$ and $\phi_1,\ldots,\phi_N$.
\end{defi}
\begin{defi}
 A distribution is a \emph{conjugate} of another distribution when they are of the same variable and of the same family.
 The two distributions are then \emph{conjugates}.
\end{defi}
\begin{ex}
 Two Gaussian distributions
 $p(x)=\N(0,1)$ and $p(y)=\N(2,3)$ are of the same family (0,1,2,3 are the parameters).
 $p(x)$ and $q(x)=\N(4,5)$ are conjugates.
 $p(x)$ and $p(x|z)=\N(2z+1,1)$ are also conjugates.
\end{ex}
\begin{conv}
 Let $x$ be an observable RV and $z$ be a latent RV.
 When a prior distribution $p(z)$ is a conjugate \uline{of} a posterior distribution $p(z|x)$,
 $p(z)$ is a \emph{conjugate prior distribution} \uline{for}
 a generative distribution $p(x|z)$.
\end{conv}

\begin{conv}
 A prior distribution is \emph{informative} if its parameters are chosen by
 the domain knowledge.
\end{conv}

\begin{conv}
 A prior distribution is \emph{non-informative} if its parameters are selected
 by the principle of maximum entropy due to the lack of such a domain knowledge.
\end{conv}

\begin{conv}
 A non-informative prior is \emph{improper} if
 the prior does not satisfy the probability axiom at the limit of entropy maximization.
\end{conv}

In the following sections, we follow a fixed format shown below to prove each conjugate prior.

\begin{conv}[Bayesian Reasoning with a single unknown parameter]
 \label{conv:bayesian-reasoning}
 Bayesian reasoning is a form of Bayesian statistical modeling \cite{gelman1995bayesian} applied as follows:
 \begin{enumerate}
  \item List observable RVs: $X=(x_1,\ldots,x_n)$.
  \item Latent RVs: A parameter $\theta$.
  \item Causal dependency:
        Assume each observation is i.i.d. given $\theta$, i.e.,
        $x_i\perp x_j \mid \theta$ and $p(x_i|\theta)=p(x_j|\theta)$.
        In other words, $p(X)=\sum_{\theta} p(\theta)\prod_i p(x_i|\theta)$.
  \item Choose a distribution family and its parameters.
        \begin{enumerate}
         \item Choose a family for $p(x_i|\theta)$.
         \item Write down $p(x_i|\theta)$.
         \item Write down $p(X|\theta) = p(x_1,\ldots,x_n|\theta) = \prod_i p(x_i|\theta)$.
         \item Perform a dimensional analysis on $\theta$ and $n$ using
               $p(X)=\frac{p(X|\theta)p(\theta)}{p(\theta|X)}=f(\theta)\frac{p(\theta)}{p(\theta|X)}$
               being constant for $\theta$.
               Choose a family and parameters for $p(\theta)$ and $p(\theta|X)$ based on the analysis,
               assuming $p(\theta)$ is parameterized by pseudocount $n_0$ and the prior parameter $\theta_0$,
               as well as $p(\theta|X)$ by pseudocount $n+n_0$ and the updated parameter $\theta_n$.
         \item Write down $p(\theta)$.
         \item Derive $p(\theta|X)$ and the parameter $\theta_n$.
               This is often done in one of the following manners:
               \begin{enumerate}
                \item Use $p(\theta|X)=\frac{p(X|\theta)p(\theta)}{p(X)}\propto p(X|\theta)p(\theta)$ with $p(X)$ being constant.
                      Ignore constant factors and match $\theta$'s coefficients in the result with the pdf of $\theta$.
                \item Derive $p(\theta,X) = p(X|\theta)p(\theta)$,
                      then $p(X)=\int p(\theta,X)d\theta$,
                      then $p(\theta|X)=\frac{p(\theta,X)}{p(X)}$.
               \end{enumerate}
        \end{enumerate}
  \item Derive a \emph{predictive distribution} $p(x|X)$ for a future data $x$ given historical data $X$.
        This is using the fact that $x$ does not depend on $X$ given $\theta$.
        \[
        p(x|X)=\int p(x|\theta,X)p(\theta|X)d\theta=\int p(x|\theta)p(\theta|X)d\theta.
        \]
        Using held-out data $x'_1,\ldots x'_M$,
        verify the hypothesis (statistical model) made above by
        computing $p(X'|X)=\prod_i p(x=x'_i|X)$
        (or, alternatively, $\log p(X'|X)=\sum_i \log p(x=x'_i|X)$).
  \item If possible, derive an improper non-informative prior.
 \end{enumerate}
\end{conv}

Finally, we derive a multi-parameter reasoning method from single-parameter reasoning methods.
The goal of multi-parameter Bayesian reasoning is
to obtain the joint posterior distribution of the parameters $p(\theta_1\ldots,\theta_N|X)$
and subsequently obtain the predictive distribution $p(x|X)$ to test the hypothesis.
Assuming the availability of a single parameter reasoning process for each parameter of the distribution family,
one should derive $p(\theta_i|X,\theta_1,\ldots,\theta_{i-1})=\frac{p(X|\theta_1,\ldots,\theta_{i-1})p(\theta_i|\theta_1,\ldots,\theta_{i-1})}{p(X|\theta_1,\ldots,\theta_{i-1})}$
for each $i$, starting from $p(\theta_1|X)$.

The only difference between single- and multi-parameter cases is that
the process involves decomposing a more complex hierarchical model,
and that it requires a \emph{joint} prior distribution $p(\theta_1\ldots,\theta_N)$.
However, the derivation from a joint prior tends to be overly complex,
therefore we sometimes use uninformative priors for mathematical convenience,
as well as assuming the independence between priors, i.e., $p(\theta_1\ldots,\theta_N)=p(\theta_1)\ldots p(\theta_N)$.
Although uninformative priors are not particularly helpful/informational in the reasoning,
this is not an issue because
the effect/importance of the prior distributions diminishes
as the depth of the model hierarchy increases.

\begin{center}
 \emph{The proofs start after the references.}
\end{center}

\fontsize{9.5pt}{10.5pt}
\selectfont

\clearpage
\section{Analytical subset for $\xi>0$: Pareto}

For $l>0$ and $\alpha>0$,
Pareto distribution $\pareto(\alpha,l)$ is a special case of $\gp$ as follows:

\begin{align*}
 \gp\parens{l, \frac{l}{\alpha}, \frac{1}{\alpha}}
 &= \frac{\alpha}{l} (1+\frac{\frac{1}{\alpha} (x-l)}{\frac{l}{\alpha}})^{-1-\alpha}
 = \frac{\alpha}{l} \parens{\frac{x}{l}}^{-1-\alpha}\\
 &= \frac{\alpha l^{\alpha}}{x^{\alpha+1}}=\pareto(\alpha,l).
\end{align*}

\begin{figure}[h]
 \includegraphics[width=\linewidth]{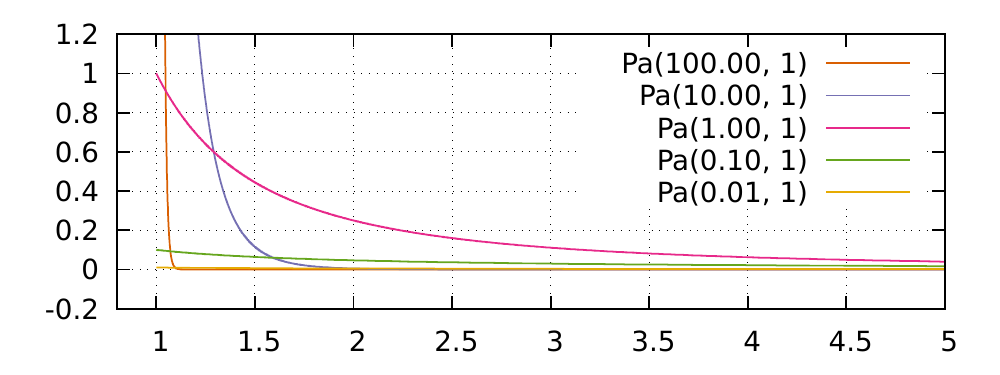}
 \caption{Pareto distribution $\pareto(\alpha,1)$.}
 \label{fig:pareto}
\end{figure}

Proofs for the conjugate priors were first given in \citet{malik1970estimation}.

\newpage
\subsection{Pareto with Unknown Lower Bound $l$}

\begin{ex}
 Checking laptop prices online,
 I found $n=20$ offers and I believe it follows $p(x_i|l)=\pareto(\alpha=1.2, l)$
 for some cheapest / minimum price $l$ that I want to know.
 I know conservatively a laptop should cost at least $l_0=100$ USD, i.e., a prior assumption.
 Can we improve $l_0$ using data?
\end{ex}

\begin{enumerate}
 \setcounter{enumi}{1}
 \item Latents: $l$.
 \setcounter{enumi}{3}
 \item Distribution family and parameters:
       \begin{enumerate}
        \item $p(x_i|l)=\pareto(x_i|\alpha,l). \quad (0<l<x_i)$.
        \item $p(x_i|l)=\alpha l^\alpha x_i^{-(\alpha+1)}$.
        \item $p(X|l)=\alpha^n l^{\alpha n} \prod_i x_i^{-(\alpha+1)}. \quad (0<l<\min_i x_i=\el)$
        \item The coefficients of $l$ in $p(l)$ and $p(l|X)$ differ by
              $l^{\alpha n}$.
              This matches Power distributions $\power(a, b)=\frac{b x^{b-1}}{a^b}$ $(0<x<a)$
              with following parameters:

              $p(l)=\power(l_0,\alpha n_0)$ ($0<l<l_0$) and

              $p(l|X)=\power(l_n,\alpha (n_0+n))$ ($0<l<l_n$).
        \item $p(l)= \alpha n_0 l_0^{-\alpha n_0}l^{\alpha n_0-1}$.
        \item Using the second strategy.
              Let $l_n=\min (l_0, \el)$.
              \begin{align*}
               p(l,X)
               &=p(X|l)p(l) \quad (0<l<l_n)\\
               &= A l^{\alpha (n+n_0)-1}. \\
               p(X)
               &\textstyle
               =\int_{0}^{l_n} p(l,X)dl
               =\frac{A}{\alpha (n+n_0)} l_n^{\alpha (n+n_0)} - 0.\\
               p(l|X)&
               =\frac{p(l,X)}{p(X)}
               =\alpha (n+n_0) l_n^{-\alpha (n+n_0)} l^{\alpha (n+n_0)-1} \\
               &
               =\power(l_n,\alpha (n_0+n)).
              \end{align*}
              In other words, the new lower bound $l_n$ updated from the prior lower bound $l_0$
              is the minimum of $l_0$ and the empirical minimum $\epi$.
       \end{enumerate}
 \item $p(x|X)=\int_{0}^{l_n} p(x|l)p(l|X)dl$
       \begin{align*}
        &=\int_{0}^{l_n} \alpha x^{-(\alpha+1)} \alpha (n+n_0) l_n^{-\alpha (n+n_0)} l^{\alpha (n+n_0+1)-1} dl\\
        &=\brackets{\alpha x^{-(\alpha+1)} \alpha (n+n_0) l_n^{-\alpha (n+n_0)} \frac{l^{\alpha (n+n_0+1)}}{\alpha (n+n_0+1)}}_{0}^{l_n}\\
        &=\alpha \frac{n+n_0}{n+n_0+1} l_n^{\alpha} x^{-(\alpha+1)}
        = \pareto\parens{x\mid \alpha, \parens{\frac{n+n_0}{n+n_0+1}}^{\frac{1}{\alpha}}l_n}.
       \end{align*}
       Notice that the new lower bound $\parens{\frac{n+n_0}{n+n_0+1}}^{\frac{1}{\alpha}}l_n$
       is smaller than the posterior lower bound $l_n$ or the empirical lower bound $\el$
       because $\parens{\frac{n+n_0}{n+n_0+1}}^{\frac{1}{\alpha}}<1$.
       Bayesian reasoning thus allows extrapolation from the data.
 \item A non-informative prior is obtained by $n_0\to \alpha^{-1}$:
       \begin{align*}
        p(l)&\to l_0^{-1}: \text{Const}.\\
        p(l|X)&\to \power(l_n,\alpha n + 1).
       \end{align*}
\end{enumerate}

\newpage
\subsection{Pareto with Unknown Shape $\alpha$}

\begin{ex}
 Checking laptop prices online,
 I found $n=20$ offers which follow $p(x_i|\alpha)=\pareto(\alpha, l=100)$.
 I want to know $\alpha$ which tells the variability.
 I have a a prior assumption $p(\alpha)=\Gamma(2, 2)$ (Gamma distribution).
\end{ex}

\begin{enumerate}
 \setcounter{enumi}{1}
 \item Latents: $\alpha$.
 \setcounter{enumi}{3}
 \item Distribution family and parameters:
       \begin{enumerate}
        \item $p(x_i|\alpha)=\pareto(x_i|\alpha,l). \quad (0<\alpha)$
        \item $p(x_i|\alpha)=\alpha l^\alpha x_i^{-(\alpha+1)}$.
        \item $p(X|\alpha)=\alpha^n l^{\alpha n} \prod_i x_i^{-(\alpha+1)} \quad (0<\alpha)$

              $=\alpha^n l^{\alpha n} l^{-n(\alpha+1)} \prod_i \parens{\frac{x_i}{l}}^{-(\alpha+1)}$

              $=\alpha^n l^{-n} \prod_i \parens{\frac{x_i}{l}}^{-(\alpha+1)}.$

              Let the geometric mean relative to $l$ be $\eg=\prod_i \parens{\frac{x_i}{l}}^{\frac{1}{n}}$.

              Then
              $p(X|\alpha)= \alpha^n l^{-n} \eg^{-n(\alpha+1)}\propto \alpha^n e^{-n \log \eg \alpha}$.

        \item The coefficients of $\alpha$ in $p(\alpha)$ and $p(\alpha|X)$ differ by
              $\alpha^n e^{-n \log \eg \alpha}$.
              This matches Gamma distributions $\Gamma(a, b)\propto x^{a-1}e^{-bx}$
              with following parameters:

              $p(\alpha)=\Gamma(n_0, n_0\log g_0)$ and

              $p(\alpha|X)=\Gamma(n_0+n, (n+n_0)\log g_n)$.
        \item $p(\alpha)\propto \alpha^{n_0-1}e^{-n_0 \log g_0\alpha}=\alpha^{n_0-1}g_0^{-n_0\alpha}$.
        \item Using the first strategy.
              $p(\alpha,X)=p(X|\alpha)p(\alpha)\propto$
              \begin{align*}
               \alpha^{(n+n_0)-1} \parens{g_0^{n_0} \eg^n}^{-\alpha}
               &\propto \alpha^{(n+n_0)-1} \parens{g_n}^{-(n+n_0)\alpha} \\
               &\propto p(\alpha|x),
              \end{align*}
              where $g_n=\parens{g_0^{n_0}\eg^n}^{\frac{1}{n+n_0}}$.
              In other words, the new geometric mean $g_n$ relative to $l$
              is a geometric mean of $g_0$ weighted by $n_0$ and
              the empirical geometric mean $\eg$ weighted by $n$.

              Let $A=n_0+n, B=(n+n_0)\log g_n$.
       \end{enumerate}
 \setcounter{enumi}{5}

 \setcounter{enumi}{4}
 \item Let $p(\alpha|X)=\Gamma(A,B)$.
       \begin{align*}
        p(x,\alpha|X)
        &\textstyle = p(x|\alpha)p(\alpha|X)\\
        &\textstyle = \alpha l^\alpha x^{-(\alpha+1)} \cdot \frac{B^A}{\Gamma(A)}\alpha^{A-1} e^{-B\alpha}\\
        &\textstyle = \frac{B^A}{x\Gamma(A)}\alpha^{A} e^{-(B+\log \frac{x}{l})\alpha}.\\
        p(x|X)
        &\textstyle = \int_0^\infty p(x,\alpha|X) d\alpha \\
        &\textstyle = \frac{B^A}{x\Gamma(A)}\frac{\Gamma(A+1)}{(B+\log \frac{x}{l})^{A+1}} \quad (\because \int_0^\infty x^ae^{-bx}dx=\frac{\Gamma(a+1)}{b^{a+1}})\\
        &\textstyle = \frac{AB^A}{x(B+\log \frac{x}{l})^{A+1}}                             \quad (\because \Gamma(z+1)=z\Gamma(z))\\
        &\textstyle = \frac{dy}{dx}\frac{AB^A}{(B+y)^{A+1}} = \frac{dy}{dx}\pareto(y-B|A,B).  \quad (y=\log \frac{x}{l})\\
        &\textstyle\therefore p(\log\frac{x}{l}+B|X)=\pareto(A,B).\\
       \end{align*}

 \item A non-informative improper prior distribution is obtained by the limit of $n_0\downto 0$:
       \begin{align*}
        p(\alpha)= \alpha^{n_0-1}\parens{\frac{l}{l_0}}^{n_0\alpha} &\to[n_0\downto 0] \alpha^{-1}.\\
        p(\alpha|X) &\to[n_0\downto 0] \Gamma(n, n\log \frac{\epi}{l}).
       \end{align*}
\end{enumerate}

\newpage
\subsection{Pareto with Unknown $\alpha,l$}

\begin{enumerate}
 \setcounter{enumi}{1}
 \item Latents: $l,\alpha$.
 \setcounter{enumi}{3}
 \item Distribution family and parameters:
       \begin{enumerate}
        \item $p(x_i|l,\alpha)=\pareto(x_i|\alpha,l). \quad (0<l<x_i)$.
        \item $p(x_i|l,\alpha)=\alpha l^\alpha x_i^{-(\alpha+1)}$.
        \item $p(X|l,\alpha)=\alpha^n l^{\alpha n} \prod_i x_i^{-(\alpha+1)}$.

              Let the geometric mean of data be $\eg = \prod_i x_i^{\frac{1}{n}}$.
              Then
              $p(X|l,\alpha)= \alpha^n l^{\alpha n} \eg^{-n(\alpha+1)}\propto \alpha^n l^{\alpha n} e^{-n \log \eg \alpha}$.

        \item $\begin{aligned}[t]
               p(l,\alpha)&=p(l|\alpha)p(\alpha)\\
               &=\power(l|l_0, \alpha n_0)\cdot \Gamma(\alpha|n'_0, n'_0\log g_0).
              \end{aligned}$

        \setcounter{enumii}{5}
        \item Using the first strategy.

              $p(l|\alpha,X) p(\alpha|X)=p(l,\alpha|X)\propto p(X|l,\alpha)p(l|\alpha)p(\alpha)$
              \begin{align*}
               &\propto \alpha^n l^{n\alpha} e^{-n\log \eg \alpha} p(l|\alpha)p(\alpha)\\
               &\propto l^{n\alpha} p(l|\alpha) \cdot \alpha^{n} e^{-n\log \eg \alpha} p(\alpha)\\
               &\propto \power(l|l_n,(n+n_0)\alpha) \cdot \Gamma(\alpha|n+n'_0, (n+n'_0)\log g_n)
              \end{align*}
              where $l_n=\min(l_0,\el)$, $g_n=\parens{g_0^{n'_0}\eg^n}^{\frac{1}{n+n'_0}}$.

              Let $A=n'_0+n, B=(n+n'_0)\log g_n$.
       \end{enumerate}
 \item $p(x|X)=\int p(x,l,\alpha|X)dld\alpha.$
       \begin{align*}
        &p(x,l,\alpha|X)=p(x|l,\alpha)p(l|\alpha,X)p(\alpha|X)\\
        &= \frac{\alpha l^\alpha}{x^{\alpha+1}} \cdot \frac{\alpha(n+n_0)l^{\alpha(n+n_0)-1}}{l_n^{\alpha(n+n_0)}} \cdot \frac{B^A}{\Gamma(A)} \alpha^{A-1}e^{-B\alpha}\\
        &= \frac{\alpha(n+n_0)l^{\alpha(n+n_0+1)-1}}{l_n^{\alpha(n+n_0)}}\cdot \frac{B^A}{x\Gamma(A)} \alpha^{A}e^{-(B+\log x)\alpha}.\\
        &p(x,\alpha|X)=\int_0^{l_n} p(x,l,\alpha|X)dl \\
        &= \frac{\alpha(n+n_0)}{l_n^{\alpha(n+n_0)}} \frac{l_n^{\alpha(n+n_0+1)}-0}{\alpha(n+n_0+1)}\cdot \frac{B^A}{x\Gamma(A)} \alpha^{A}e^{-(B+\log x)\alpha}\\
        &= \frac{n+n_0}{n+n_0+1} \cdot \frac{B^A}{x\Gamma(A)} \alpha^{A}e^{-(B+\log \frac{x}{l_n})\alpha}.\\
        &p(x|X)\textstyle = \int_{0}^\infty p(x,\alpha|X)d\alpha \\
        &\textstyle = \frac{n+n_0}{n+n_0+1} \cdot \frac{B^A}{x\Gamma(A)}\frac{\Gamma(A+1)}{(B+\log \frac{x}{l_n})^{A+1}} \quad (\because \int_0^\infty x^ae^{-bx}dx=\frac{\Gamma(a+1)}{b^{a+1}})\\
        &\textstyle = \frac{n+n_0}{n+n_0+1} \cdot \frac{AB^A}{x(B+\log \frac{x}{l_n})^{A+1}}                             \quad (\because \Gamma(z+1)=z\Gamma(z))\\
        &\textstyle = \frac{n+n_0}{n+n_0+1} \frac{dy}{dx}\frac{AB^A}{y^{A+1}}. \quad (y=\log \frac{x}{l_n}+B)\\
        &\textstyle \therefore p(\log\frac{x}{l_n}+B|X)= \pareto(A,\parens{\frac{n+n_0}{n+n_0+1}}^{\frac{1}{A}}B).
       \end{align*}
       We observe a Bayesian extrapolation similar to the single-parameter case.
       From the lower bound of the Pareto distribution,
       \begin{align*}
        y=\log \frac{x}{l_n} + B\in \left[\parens{\frac{n+n_0}{n+n_0+1}}^{\frac{1}{A}}B,\quad \infty\right]&=[y^{-*},\infty]\\
        x\in \left[l_n\exp B\parens{\parens{\frac{n+n_0}{n+n_0+1}}^{\frac{1}{A}}-1}, \infty\right]&=[x^{-*},\infty]
       \end{align*}
       Since $\parens{\frac{n+n_0}{n+n_0+1}}^{\frac{1}{A}}-1<0$, this lower bound $x^{-*}$ is smaller than $l_n$ and $\el$.
\end{enumerate}

\clearpage
\section{Analytical subset for $\xi=0$: Shifted Exponential}

For $l>0$ and $\alpha>0$,
a shifted Exponential distribution $\Exp(\alpha,l)$ is a special case of $\gp$ as follows:

\begin{align*}
 \gp\parens{l, \frac{1}{\alpha}, 0}
 &= \alpha \exp -\frac{x-l}{\frac{1}{\alpha}}
  = \alpha e^{-\alpha(x-l)} = \Exp(\alpha,l).
\end{align*}

\begin{figure}[h]
 \includegraphics[width=\linewidth]{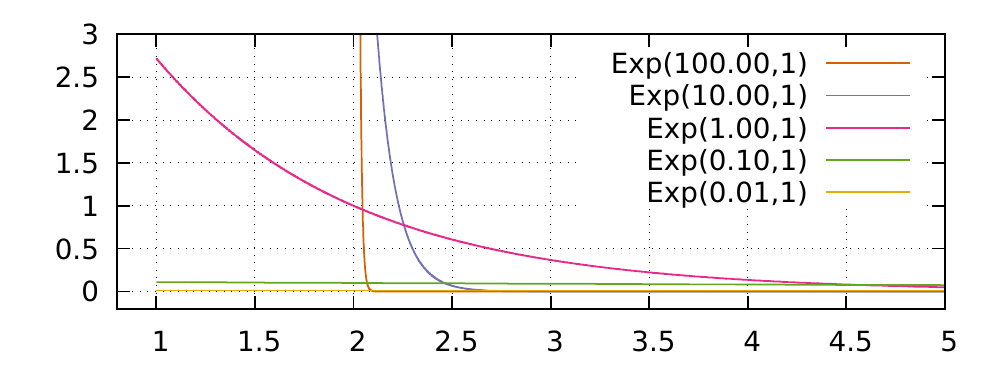}
 \caption{Shited Exponential distribution $\Exp(\alpha,1)$.}
 \label{fig:exp}
\end{figure}

\newpage
\subsection{Shifted Exponential with Unknown $l$}

\begin{enumerate}
 \setcounter{enumi}{1}
 \item Latents: $l$.
 \setcounter{enumi}{3}
 \item Distribution family and parameters:
       \begin{enumerate}
        \item $p(x_i|l)=\Exp(x_i|\alpha,l). \quad (l<x_i)$.
        \item $p(x_i|l)=\alpha e^{-\alpha(x_i-l)}$.
        \item $p(X|l)=\alpha^n e^{-\alpha n (\emu-l)} \quad (l<\min_i x_i = \el)$

              where $\emu=\frac{1}{n}\sum_i x_i$.
        \item The coefficients of $l$ in $p(l)$ and $p(l|X)$ differ by
              $e^{n\alpha l}$.
              This matches Log Power distributions $\logpower(a,b)=be^{b(x-a)}$ ($x<a$):

              $p(l)=\logpower (l_0, \alpha n_0)$ ($l<l_0$) and

              $p(l|X)=\logpower (l_n, \alpha(n+n_0))$ ($l<l_n$).

        \item $p(l)=n_0\alpha e^{\alpha n_0(l-l_0)}$.
        \item Using the second strategy.
              Let $l_n=\min (l_0, \el)$.
              \begin{align*}
               p(l,X)
               &=p(X|l)p(l)\quad (l<l_n)\\
               &=Ae^{\alpha (n+n_0) l}.\\
               p(X)
               &\textstyle
               =\int_{-\infty}^{l_n} p(l,X)dl
               =\frac{A}{\alpha (n+n_0)} e^{\alpha (n+n_0) l_n} - 0.\\
               p(l|X)&
               =\frac{p(l,X)}{p(X)}
               =\alpha (n+n_0) e^{\alpha (n+n_0) (l-l_n)}\\
               &
               =\logpower (l_n, \alpha(n+n_0)).
              \end{align*}
              In other words, the new lower bound $l_n$ updated from the prior lower bound $l_0$
              is the minimum of $l_0$ and the empirical minimum $\el$.
       \end{enumerate}
 \item $p(x|X)=\int_{-\infty}^{l_n} p(x|l)p(l|X)dl$
       \begin{align*}
        &=\int_{-\infty}^{l_n} \alpha e^{-\alpha(x-l)} \alpha (n+n_0) e^{\alpha (n+n_0) (l-l_n)}\\
        &=\int_{-\infty}^{l_n} \alpha^2 (n+n_0) e^{-\alpha (x + (n+n_0) l_n)} e^{\alpha (n+n_0+1) l}\\
        &=\alpha^2 (n+n_0) e^{-\alpha (x + (n+n_0) l_n)} \frac{e^{\alpha (n+n_0+1) l_n}}{\alpha (n+n_0+1)}-0\\
        &=\alpha \frac{n+n_0}{n+n_0+1} e^{-\alpha (x - l_n)}
        =\alpha e^{-\alpha (x - l_n - \frac{1}{\alpha}\log \frac{n+n_0}{n+n_0+1})}\\
        &= \Exp\parens{\alpha,l_n+\log \parens{\frac{n+n_0}{n+n_0+1}}^{\frac{1}{\alpha}}}.
       \end{align*}
       Notice that the new lower bound $l_n+\log \parens{\frac{n+n_0}{n+n_0+1}}^{\frac{1}{\alpha}}$
       is smaller than the posterior lower bound $l_n$ or the empirical lower bound $\el$
       because $\parens{\frac{n+n_0}{n+n_0+1}}^{\frac{1}{\alpha}}<1$.
       Bayesian reasoning thus allows extrapolation from the data.
 \item A non-informative improper prior distributions is obtained by the limit of $n_0\downto 0$:
       \begin{align*}
        p(l)\propto e^{\alpha n_0 l}
        &\to[n_0\downto 0] C: \text{Const.}\\
        p(l|X)
        &\to[n_0\downto 0] \logpower(\alpha n, \el)\\
       \end{align*}
\end{enumerate}

Note: When $y\sim \power(\alpha,\beta)$ ($0<y<\alpha$),
then $\log y=x\sim \logpower(a,b)$ ($-\infty<x<a=\log \alpha, b=\beta$).
This follows from using $y=e^x$ and $dy=e^xdx$,
\[
\hspace{-2em}
\int_0^\alpha \frac{\beta y^{\beta-1}}{\alpha^\beta}dy=
\int_{-\infty}^a \hspace{-0.5em}\frac{b e^{(b-1)x}}{e^{ab}}e^xdx=
\int_{-\infty}^a \hspace{-0.5em} b e^{b(x-a)}dx= 1.
\]

\newpage
\subsection{Shifted Exponential with Unknown $\alpha$}

\begin{enumerate}
 \setcounter{enumi}{1}
 \item Latents: $\alpha$.
 \setcounter{enumi}{3}
 \item Distribution family and parameters:
       \begin{enumerate}
        \item $p(x_i|\alpha)=\Exp(x_i|\alpha,l). \quad (0<\alpha)$.
        \item $p(x_i|\alpha)=\alpha e^{-\alpha(x_i-l)}$.
        \item $p(X|\alpha)=\alpha^n e^{-\alpha n (\emu-l)} \quad (0<\alpha)$

              where $\emu=\frac{1}{n}\sum_i x_i$.

        \item The coefficients of $\alpha$ in $p(\alpha)$ and $p(\alpha|X)$ differ by
              $\alpha^n e^{-n (\emu-l) \alpha }$.
              This matches Gamma distributions $\Gamma(a, b)\propto x^{a-1}e^{-bx}$
              with following parameters:

              $p(\alpha)=\Gamma(n_0, n_0 (\mu_0-l))$ and

              $p(\alpha|X)=\Gamma(n_0+n, (n+n_0) (\mu_n-l))$.

        \item $p(\alpha)\propto \alpha^{n_0-1}e^{-n_0 (\mu_0-l)\alpha}$.
        \item Using the first strategy.
              $p(\alpha,X)=p(X|\alpha)p(\alpha)$
              \begin{align*}
               &\textstyle \propto\alpha^{(n+n_0)-1} e^{\alpha\parens{n_0(\mu_0-l)+n(\emu-l)}}\\
               &\textstyle \propto\alpha^{(n+n_0)-1} e^{\alpha (n+n_0)(\mu_n-l)}\propto\hspace{0.7em} p(\alpha|x),
              \end{align*}
              where $\mu_n=\frac{n\emu+n_0\mu_0}{n+n_0}$.
              In other words, the new mean $\mu_n$
              is an average of $\mu_0$ and the empirical mean $\emu$ weighted by $n_0$ and $n$, respectively.

              Let $A=n_0+n, B=(n+n_0) (\mu_n-l)$.
       \end{enumerate}
 \item Let $p(\alpha|X)=\Gamma(A,B)$.
       \begin{align*}
        p(x,\alpha|X)
        &\textstyle =p(x|\alpha)p(\alpha|X)\\
        &\textstyle = \alpha e^{-\alpha(x-l)} \cdot \frac{B^A}{\Gamma(A)}\alpha^{A-1} e^{-B\alpha}\\
        &\textstyle = \frac{B^A}{\Gamma(A)}\alpha^{A} e^{-(B+x-l)\alpha}\\
        p(x|X)
        &\textstyle = \int_0^\infty p(x,\alpha|X) d\alpha \\
        &\textstyle = \frac{B^A}{\Gamma(A)}\frac{\Gamma(A+1)}{(B+x-l)^{A+1}} \quad (\because \int_0^\infty x^ae^{-bx}dx=\frac{\Gamma(a+1)}{b^{a+1}})\\
        &\textstyle = \frac{AB^A}{(B+x-l)^{A+1}}                             \quad (\because \Gamma(z+1)=z\Gamma(z))\\
        &\textstyle \therefore p(x+B-l|X)= \pareto(A,B).
       \end{align*}

 \item A non-informative improper prior distribution is obtained by the limit of $n_0\downto 0$:
       \begin{align*}
        p(\alpha)= \alpha^{n_0-1}\parens{\frac{l}{l_0}}^{n_0\alpha} &\to[n_0\downto 0] \alpha^{-1}.\\
        p(\alpha|X) &\to[n_0\downto 0] \Gamma(n, n(\emu-l)).
       \end{align*}
\end{enumerate}

\newpage
\subsection{Shifted Exponential with Unknown $\alpha,l$}

\begin{enumerate}
 \setcounter{enumi}{1}
 \item Latents: $l,\alpha$.
 \setcounter{enumi}{3}
 \item Distribution family and parameters:
       \begin{enumerate}
        \item $p(x_i|l,\alpha)=\Exp(x_i|\alpha,l).$.
        \item $p(x_i|l,\alpha)=\alpha e^{-\alpha(x_i-l)}$.
        \item $p(X|l,\alpha)=\alpha^n e^{-\alpha n (\emu-l)}.$
        \item $\begin{aligned}[t]
               p(l,\alpha)&=p(l|\alpha)p(\alpha)\\
               &=\logpower(l|l_0, \alpha n_0)\cdot \Gamma(\alpha|n'_0, n'_0 \mu_0).
              \end{aligned}$

        \setcounter{enumii}{5}
        \item Using the first strategy.

              $p(l|X,\alpha)p(\alpha|X)=p(l,\alpha|X)\propto p(X|l,\alpha)p(l|\alpha)p(\alpha)$
              \begin{align*}
               &\propto \alpha^n e^{-\alpha n (\emu-l)} \cdot p(l|\alpha)p(\alpha)\\
               &\propto e^{\alpha nl} p(l|\alpha) \cdot \alpha^n e^{- n \emu \alpha} p(\alpha)\\
               &\propto \logpower(l|l_n,\alpha(n+n_0)) \cdot \Gamma(\alpha|n+n'_0,(n+n'_0) \mu_n)
              \end{align*}
              where $l_n=\min(l_0,\el)$, $\mu_n=\frac{n'_0\mu_0+n\emu}{n'_0+n}$.

              Let $A=n'_0+n, B=(n+n'_0) \mu_n$.
       \end{enumerate}
 \item $p(x|X)=\int p(x,l,\alpha|X)dld\alpha$
       \begin{align*}
        &p(x,l,\alpha|X)=p(x|l,\alpha)p(l|\alpha,X)p(\alpha|X)\\
        &= \alpha e^{-\alpha (x-l)} \cdot \alpha(n+n_0)e^{\alpha(n+n_0)(l-l_n)} \cdot \frac{B^A}{\Gamma(A)} \alpha^{A-1}e^{-B\alpha}\\
        &= \alpha(n+n_0) e^{\alpha(n+n_0+1)l} \cdot \frac{B^A}{\Gamma(A)} \alpha^{A}e^{-(B+x+(n+n_0)l_n)\alpha}.\\
        &p(x,\alpha|X)=\int_{-\infty}^{l_n} p(x,l,\alpha|X)dl \\
        &= \alpha(n+n_0) \frac{e^{\alpha(n+n_0+1)l_n}-0}{\alpha(n+n_0+1)}  \cdot \frac{B^A}{\Gamma(A)} \alpha^{A}e^{-(B+x+(n+n_0)l_n)\alpha}\\
        &= \frac{n+n_0}{n+n_0+1} \frac{B^A}{\Gamma(A)} \alpha^{A}e^{-(B+x-l_n)\alpha}.\\
        &p(x|X)=\int_{0}^\infty p(x,\alpha|X)d\alpha \\
        &\textstyle = \frac{n+n_0}{n+n_0+1} \cdot \frac{B^A}{\Gamma(A)}\frac{\Gamma(A+1)}{(B+x-l_n)^{A+1}} \quad (\because \int_0^\infty x^ae^{-bx}dx=\frac{\Gamma(a+1)}{b^{a+1}})\\
        &\textstyle = \frac{n+n_0}{n+n_0+1} \cdot \frac{AB^A}{(B+x-l_n)^{A+1}}                             \quad (\because \Gamma(z+1)=z\Gamma(z))\\
        &\textstyle \therefore p(x+B-l_n|X)= \pareto(A,\parens{\frac{n+n_0}{n+n_0+1}}^{\frac{1}{A}}B).
       \end{align*}
       We observe a Bayesian extrapolation similar to the single-parameter case.
       From the lower bound of the Pareto distribution,
       \begin{align*}
        y=x+B-l_n\in \left[\parens{\frac{n+n_0}{n+n_0+1}}^{\frac{1}{A}}B,\infty\right] &=[y^{-*},\infty] \\
        x\in \left[l_n + B\parens{\parens{\frac{n+n_0}{n+n_0+1}}^{\frac{1}{A}}-1},\infty\right] &=[x^{-*},\infty].
       \end{align*}
       Since $\parens{\frac{n+n_0}{n+n_0+1}}^{\frac{1}{A}}-1<0$, this lower bound $x^{-*}$ is smaller than $l_n$ and $\el$.
\end{enumerate}

\clearpage
\section{Analytical subset for $\xi<0$: Reverted Power (or: Inverted Pareto)}

When a RV $y$ follows a distribution $f$,
its inverse $x=\frac{1}{y}$ is said to follow an inverted distribution of $f$.
An Inverted Pareto $\frac{1}{y}=x\sim\ipareto(\alpha,l)$ is thus defined from $y\sim\pareto(\alpha,l)$.
\begin{align*}
 &1
 = \int_{l}^{\infty} \pareto(y\mid\alpha,l)dy
 = \int_{l}^{\infty} \frac{\alpha l^\alpha}{y^{\alpha+1}}dy\\
 &
 = \int_{l^{-1}}^{0} \hspace{-0.7em}\alpha l^\alpha x^{\alpha+1} \parens{\frac{-1}{x^2}dx}
 = \int_{0}^{l^{-1}} \hspace{-0.7em}\alpha l^\alpha x^{\alpha-1} dx.\\
 &
 \alpha l^\alpha x^{\alpha-1} = \ipareto(\alpha,l) = \power(l^{-1},\alpha). \quad (0<x<l^{-1})
\end{align*}

Inverted Pareto is equivalent to a Power distribution \cite{dallas1976characterizing},
which is defined as follows:
\[
 \power(a,b)=\frac{bx^{b-1}}{a^b}.\quad (0<x<a)
\]

When a RV $y$ follows a distribution $f$,
$x=-y$ is said to follow a reverted distribution of $f$.
Inverted Pareto / Power is a reverted distribution of a special case of Generalized Pareto as follows:
\begin{align*}
 \gp\parens{-a,\frac{a}{b},-\frac{1}{b}}
 &=\frac{b}{a}\parens{1+\parens{-\frac{1}{b}}\frac{b}{a}(x-(-a))}^{-1-\frac{1}{-\frac{1}{b}}}\\
 &
 =\frac{b(-x)^{b-1}}{a^b} \\
 &=\power(-x\mid a,b)=\ipareto(-x\mid b,a^{-1}).
\end{align*}

Finally, a special case of Uniform distribution with a lower support $l=0$, i.e., $U(0,u)=\frac{1}{u}$ $(0<x<u)$,
is a special case of Power distribution $\power(a=u,b=1)$.
Note that generally the Uniform distribution is \emph{not} a special case of Power distribution.
Both Uniform and Power have two parameters, and they are obtained by reducing the degree of freedom
of the three parameter GP distribution by 1.

\begin{figure}[h]
 \includegraphics[width=\linewidth]{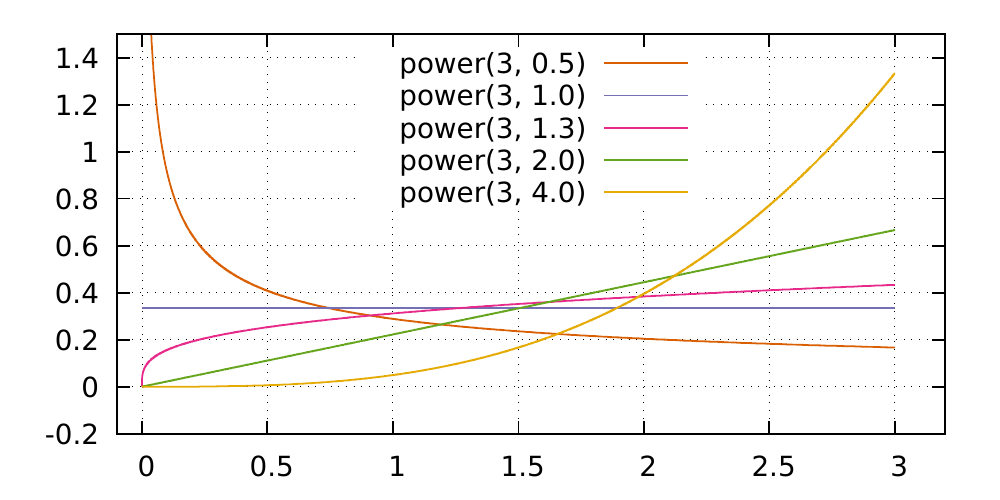}
 \caption{Power distribution $\power(3,\alpha)$.}
 \label{fig:exp}
\end{figure}

To avoid the confusion due to the inverse and the reverse,
we focus on a Power distribution in the following.

\newpage
\subsection{Power with Unknown Upper Bound $u$}

\begin{enumerate}
 \setcounter{enumi}{1}
 \item Latents: $u$.
 \setcounter{enumi}{3}
 \item Distribution family and parameters:
       \begin{enumerate}
        \item $p(x_i|u)=\power(x_i|u,\alpha). \quad (0<x_i<u)$
        \item $p(x_i|u)=\alpha u^{-\alpha} x_i^{\alpha-1}$.
        \item $p(X|u)=\alpha^n u^{-n\alpha} \prod_i x_i^{\alpha-1}. \quad (0<\max_i x_i<u)$

              Let $\max_i x_i=\eu$.
        \item The coefficients of $u$ in $p(u)$ and $p(u|X)$ differ by
              $u^{-n\alpha}$.
              This matches Pareto distributions $\pareto(a, b)=\frac{a b^a}{x^{a+1}}$ $(a<x)$
              with following parameters:

              $p(u)=\pareto(\alpha n_0, u_0)$ and

              $p(u|X)=\pareto(\alpha (n+n_0), u_n)$.
        \item $p(u)=\alpha n_0 u_0^{\alpha n_0} u^{-(\alpha n_0+1)}$.
        \item Using the second strategy.
              Let $u_n=\max (u_0,\eu)$.
              \begin{align*}
               p(u,X)
               &=p(X|u)p(u) \quad (u_n < u)\\
               &=A u^{-\alpha(n+n_0)-1}.\\
               p(X)
               &\textstyle
               =\int_{u_n}^{\infty} p(u,X)dl
               =A\frac{0 - u_n^{-\alpha (n+n_0)}}{-\alpha (n+n_0)}.\\
               p(u|X)&
               =\frac{p(u,X)}{p(X)}
               =\frac{\alpha (n+n_0) u_n^{\alpha (n+n_0)}}{u^{\alpha(n+n_0)+1}} \\
               &
               =\pareto(\alpha (n+n_0), u_n).
              \end{align*}
              In other words, the new upper bound $u_n$ updated from the prior upper bound $u_0$
              is the maximum of $u_0$ and the empirical maximum $\eu$.
       \end{enumerate}
 \item $p(x|X)=\int_{u_n}^{\infty} p(x|u)p(u|X)du$
       \begin{align*}
        &=\textstyle\int_{u_n}^{\infty} \alpha u^{-\alpha} x^{\alpha-1} \cdot \frac{\alpha (n+n_0) u_n^{\alpha (n+n_0)}}{u^{\alpha(n+n_0)+1}}\\
        &=\textstyle\int_{u_n}^{\infty} \alpha x^{\alpha-1} \alpha (n+n_0) u_n^{\alpha (n+n_0)} u^{-\alpha(n+n_0+1)-1}\\
        &=\textstyle\alpha x^{\alpha-1} \alpha (n+n_0) u_n^{\alpha (n+n_0)} (0 - \frac{u_n^{-\alpha(n+n_0+1)}}{-\alpha(n+n_0+1)})\\
        &=\alpha x^{\alpha-1} \frac{n+n_0}{n+n_0+1}u_n^{-\alpha}\\
        &=\power\parens{x\mid \parens{\frac{n+n_0}{n+n_0+1}}^\frac{-1}{\alpha} u_n, \alpha}.
       \end{align*}
       Notice that the new upper bound $\parens{\frac{n+n_0}{n+n_0+1}}^\frac{-1}{\alpha} u_n$
       is larger than the posterior upper bound $u_n$ or the empirical upper bound $\eu$
       because $\parens{\frac{n+n_0}{n+n_0+1}}^{\frac{1}{\alpha}}<1$.
       Bayesian reasoning thus allows extrapolation from the data.
 \item A non-informative improper prior is obtained by the limit of
       $n_0\to 0$:
       \begin{align*}
        p(u)
        \propto u^{-\alpha n_0-1}
        &\to u^{-1}.\\
        p(u|X)
        &\to \pareto(\alpha n, \eu)\\
       \end{align*}
\end{enumerate}

\newpage
\subsection{Power with Unknown Rate $\alpha$}

\begin{enumerate}
 \setcounter{enumi}{1}
 \item Latents: $\alpha$.
 \setcounter{enumi}{3}
 \item Distribution family and parameters:
       \begin{enumerate}
        \item $p(x_i|\alpha)=\power(x_i|u,\alpha). \quad (\alpha>-1)$
        \item $p(x_i|\alpha)=\alpha u^{-\alpha} x_i^{\alpha-1}$.
        \item $p(X|\alpha)=\alpha^n u^{-n\alpha} \prod_i x_i^{\alpha-1}. \quad (\alpha>-1)$

              $=\alpha^n u^{-n\alpha} u^{n(\alpha-1)}\prod_i \parens{\frac{x_i}{u}}^{\alpha-1}$

              $=\alpha^n u^{-n} \prod_i \parens{\frac{x_i}{u}}^{\alpha-1}$.

              Let the geometric mean of the data relative to $u$ be $\eg=\prod_i \parens{\frac{x_i}{u}}^{\frac{1}{n}}$.
              Then
              $p(X|\alpha)\propto \alpha^n \eg^{n(\alpha-1)}\propto \alpha^n e^{n\log \eg \alpha}$.

        \item The coefficients of $\alpha$ in $p(\alpha)$ and $p(\alpha|X)$ differ by
              $\alpha^n e^{n\log \eg \alpha}$.
              This matches Gamma distributions $\Gamma(a, b)\propto x^{a-1}e^{-bx}$
              with following parameters:

              $p(\alpha)=\Gamma(n_0, -n_0\log g_0)$ and

              $p(\alpha|X)=\Gamma(n_0+n, -(n+n_0)\log g_n)$.
        \item $p(\alpha)\propto \alpha^{n_0-1}e^{n_0 \log g_0\alpha}=\alpha^{n_0-1}g_0^{n_0\alpha}$.
        \item Using the first strategy.
              $p(\alpha,X)=p(X|\alpha)p(\alpha)\propto$
              $\alpha^{(n+n_0)-1} \parens{g_0^{n_0}\eg^{n}}^{\alpha}\propto \alpha^{(n+n_0)-1} g_n^{(n+n_0)\alpha}\hspace{-2.2em}\propto\hspace{0.7em} p(\alpha|x)$,
              where $g_n=\parens{g_0^{n_0}\eg^n}^{\frac{1}{n+n_0}}$.
              In other words, the new rate relative to $u$
              is a geometric mean of $g_0$ weighted by $n_0$ and
              the empirical geometric mean $\eg$ weighted by $n$.
       \end{enumerate}
 \item Let $p(\alpha|X)=\Gamma(A, B)$. $B>0$ because $g_n<1$.
       \begin{align*}
        p(x,\alpha|X)
        &\textstyle = p(x|\alpha)p(\alpha|X)\\
        &\textstyle = \alpha u^{-\alpha} x^{\alpha-1} \cdot \frac{B^A}{\Gamma(A)}\alpha^{A-1} e^{-B\alpha}\\
        &\textstyle = \frac{B^A}{x\Gamma(A)}\alpha^{A} e^{-(B+\log \frac{u}{x})\alpha}\\
        p(x|X)
        &\textstyle = \int_0^\infty p(x,\alpha|X) d\alpha \\
        &\textstyle = \frac{B^A}{x\Gamma(A)}\frac{\Gamma(A+1)}{(B+\log \frac{u}{x})^{A+1}} \quad (\because \int_0^\infty x^ae^{-bx}dx=\frac{\Gamma(a+1)}{b^{a+1}})\\
        &\textstyle = \frac{AB^A}{x(B+\log \frac{u}{x})^{A+1}}                             \quad (\because \Gamma(z+1)=z\Gamma(z))\\
        &\textstyle = \frac{dy}{dx}\frac{AB^A}{(B-y)^{A+1}} = \frac{dy}{dx}\lomax(-y|A,B). \quad (y=\log \frac{x}{u})\\
        &\textstyle\therefore p(-\log\frac{x}{u}+B|X)=\pareto(A,B).
       \end{align*}
 \item A non-informative improper prior distributions is obtained by the limit of $n_0\downto 0$:
       \begin{align*}
        p(\alpha)= \alpha^{n_0-1}\parens{\frac{u_0}{u}}^{n_0\alpha} &\to[n_0\downto 0] \alpha^{-1}.\\
        p(\alpha|X) &\to[n_0\downto 0] \Gamma(n, n\log \frac{u}{\eg}).
       \end{align*}
\end{enumerate}

\newpage
\subsection{Power with Unknown $\alpha,u$}

\begin{enumerate}
 \setcounter{enumi}{1}
 \item Latents: $u,\alpha$.
 \setcounter{enumi}{3}
 \item Distribution family and parameters:
       \begin{enumerate}
        \item $p(x_i|u,\alpha)=\power(x_i|u,\alpha).$
        \item $p(x_i|u,\alpha)=\alpha u^{-\alpha} x_i^{\alpha-1}$.
        \item $p(X|u,\alpha)=\alpha^n u^{-\alpha n} \prod_i x_i^{\alpha-1}$.

              Let the geometric mean of data be $\eg = \prod_i x_i^{\frac{1}{n}}$.
              Then

              $p(X|u,\alpha)=\alpha^n u^{-n\alpha} \eg^{n(\alpha-1)}\propto\alpha^n u^{-n\alpha} e^{n\log \eg \alpha}.$
        \item $\begin{aligned}[t]
               p(u,\alpha)&=p(u|\alpha)p(\alpha)\\
               &=\pareto(u|\alpha n_0, u_0)\cdot \Gamma(\alpha|n'_0, -n'_0\log g_0)\\
              \end{aligned}$

        \setcounter{enumii}{5}
        \item $p(u|X,\alpha) p(\alpha|x)=p(u,\alpha|X)\propto p(X|u,\alpha)p(u|\alpha)p(\alpha)$
              \begin{align*}
               &\propto \alpha^n e^{n\log \eg \alpha} u^{-n\alpha} \cdot p(u|\alpha)p(\alpha)\\
               &\propto u^{-n\alpha} p(u|\alpha) \cdot \alpha^{n} e^{n\log \eg \alpha} p(\alpha) \\
               &\propto \pareto(u|\alpha(n+n_0),u_n) \cdot \Gamma(\alpha|n+n'_0, -(n+n'_0) \log g_n),
              \end{align*}
              where $u_n =\max(u_0, \max_i x_i)$, $g_n=(g_0^{n_0}\eg^n)^{\frac{1}{n+n_0}}.$

              Let $A=n+n'_0, B=-(n+n'_0) \log g_n$.
       \end{enumerate}
 \item $p(x|X)=\int p(x,u,\alpha|X)dud\alpha$
       \begin{align*}
        &p(x,u,\alpha|X)=p(x|u,\alpha)p(u|\alpha,X)p(\alpha|X)\\
        &= \frac{\alpha x^{\alpha-1}}{u^\alpha} \cdot \frac{\alpha(n+n_0)u_n^{\alpha(n+n_0)}}{u^{\alpha(n+n_0)+1}} \cdot \frac{B^A}{\Gamma(A)} \alpha^{A-1}e^{-B\alpha}\\
        &= \frac{\alpha(n+n_0)u_n^{\alpha(n+n_0)}}{u^{\alpha(n+n_0+1)+1}}\cdot \frac{B^A}{x\Gamma(A)} \alpha^{A}e^{-(B-\log x)\alpha}\\
        &p(x,\alpha|X)=\int_{u_n}^\infty p(x,u,\alpha|X)du \\
        &= \alpha(n+n_0)u_n^{\alpha(n+n_0)} \frac{0-u_n^{-\alpha(n+n_0+1)}}{-\alpha(n+n_0+1)}\cdot \frac{B^A}{x\Gamma(A)} \alpha^{A}e^{-(B-\log x)\alpha}\\
        &= \frac{n+n_0}{n+n_0+1} \cdot \frac{B^A}{x\Gamma(A)} \alpha^{A}e^{-(B+\log \frac{u_n}{x})\alpha}\\
        &p(x|X)=\int_{0}^\infty p(x,\alpha|X)d\alpha \\
        &\textstyle = \frac{n+n_0}{n+n_0+1}\frac{B^A}{x\Gamma(A)}\frac{\Gamma(A+1)}{(B+\log \frac{u_n}{x})^{A+1}} \quad (\because \int_0^\infty x^ae^{-bx}dx=\frac{\Gamma(a+1)}{b^{a+1}})\\
        &\textstyle = \frac{n+n_0}{n+n_0+1}\frac{AB^A}{x(B+\log \frac{u_n}{x})^{A+1}}                             \quad (\because \Gamma(z+1)=z\Gamma(z))\\
        &\textstyle = \frac{n+n_0}{n+n_0+1}\frac{-dy}{dx}\frac{AB^A}{y^{A+1}}.   \quad (y=\log \frac{u_n}{x}+B)\\
        &\textstyle \therefore p(\log\frac{u_n}{x}+B|X)= \pareto(A,\parens{\frac{n+n_0}{n+n_0+1}}^{\frac{1}{A}}B).
       \end{align*}
       The support is
       \[
        \begin{array}{cl}
         y=\log \frac{u_n}{x} + B \in \left[\parens{\frac{n+n_0}{n+n_0+1}}^{\frac{1}{A}}B,\quad \infty\right]&=[y^{-*},\infty]\\
         x\in \left[0,\quad u_n\exp -B\parens{\parens{\frac{n+n_0}{n+n_0+1}}^{\frac{1}{A}}-1}\right]&=[0,x^*].
        \end{array}
       \]
       The expected value of $x$ is $\E_{p(x|X)}[x]=$
       \begin{align*}
        &\textstyle=\frac{n+n_0}{n+n_0+1}\int_{0}^{x^*} \ldots dx
         \textstyle=\frac{n+n_0}{n+n_0+1}\int_{\infty}^{y^*} x \frac{-dy}{dx}\frac{AB^A}{y^{A+1}}dy\\
        &\textstyle=\frac{n+n_0}{n+n_0+1}\int_{y^*}^{\infty} u_ne^{B-y} \frac{AB^A}{y^{A+1}}dy\\
        &\textstyle=\frac{n+n_0}{n+n_0+1}u_n AB^Ae^B \int_{y^*}^{\infty} y^{-A-1}e^{-y} dy\\
        &\textstyle=\frac{n+n_0}{n+n_0+1}u_n AB^Ae^B \Gamma(-A,y^*). \quad (\text{Incomplete $\Gamma$ function}.)
       \end{align*}

\end{enumerate}

\newpage
\section{Analytical subset for $\xi=-1$: Uniform}

We consider the uniform distribution as a special case of $\gp$.

\begin{align*}
 \gp\parens{l,u-l,-1}
 &=\frac{1}{u-l}\parens{1+(-1)\frac{x-l}{u-l}}^{-1-\frac{1}{-1}}\\
 &=\frac{1}{u-l}.
\end{align*}

Since $\xi=-1<0$, it has a support $l < x < l-\frac{u-l}{-1} = u$.
This matches the uniform distribution.

Note that Pareto and Power reduces the degree of freedom in $\gp$ by
tying $\sigma$ to $\theta$ and $\xi$,
while Shifted Exponential and Uniform do so by
setting $\xi$ to a specific value.

\newpage
\subsection{Uniform $U(l,u=l+w)$ with Unknown Width $w$}

\begin{ex}[German Tank Problem]
The Allies have captured $N=100$ Nazi tanks each of which has a serial number painted on the side, starting from $l=1$.
Currently, the maximum number observed so far is $\eu=993$.
Assuming that the number is assigned uniformly,
how many tanks were likely produced?
\end{ex}

\begin{enumerate}
 \setcounter{enumi}{1}
 \item Latents: $w$.
 \setcounter{enumi}{3}
 \item Distribution family and parameters:
       \begin{enumerate}
        \item $p(x_i|w)=U(x_i|l,l+w). \quad (w>0)$
        \item $p(w)=\pareto(n_0, w_0)$.
        \item $p(w|X)=\pareto(n_0+n, w_n)$.
       \end{enumerate}
 \setcounter{enumi}{5}
 \item $p(x_i|w)=w^{-1}$, $l<x_i<l+w$.
 \item $p(X|w)=w^{-n}$, $l<\min_i x_i < \max_i x_i<l+w$.

       Let $\ew=\max_i x_i-l$.
 \item $p(w)=n_0u_0^{n_0} w^{-n_0-1}$ where $0<w_0<w$, otherwise 0.
 \item Using the second strategy.
       $p(w,X)=p(X|w)p(w)=n_0u_0^{n_0} w^{-n-n_0-1}$ where $0<\max (w_0, \ew)<w$,
       otherwise 0. Let $w_n=\max (w_0, \ew)$.
       \begin{align*}
        p(X)
        &
        \textstyle=\int_\R p(w,X)dw
        \textstyle=\int_{w_n}^\infty p(w,X)dw\\
        &
        \textstyle=\brackets{n_0u_0^{n_0} \frac{w^{-n-n_0}}{-n-n_0}}_{w_n}^\infty
        \textstyle=\frac{n_0u_0^{n_0}}{n+n_0} w_n^{-n-n_0}.\\
        p(w|X)
        &=\frac{p(w,X)}{p(X)}=\frac{(n+n_0)w_n^{n+n_0}}{w^{(n+n_0)+1}} = \pareto(n_0+n, w_n).
       \end{align*}
       In other words, the new max $w_n$ updated from the prior max $w_0$
       is the max of $w_0$ and the empirical max width $\ew$.
 \setcounter{enumi}{4}
 \item $p(x|w)p(w|X) = \frac{(n+n_0)w_n^{n+n_0}}{w^{(n+n_0)+2}}$.

       \begin{align*}
        p(x|X)
        &=\int_{w_n}^\infty p(x,w|X)dw = \int_{w_n}^\infty p(x|w)p(w|X)dw \\
        &=0-\frac{(n+n_0)w_n^{n+n_0}}{-(n+n_0+1)w_n^{(n+n_0)+1}}.\\
        &=\frac{n+n_0}{n+n_0+1}w_n^{-1}\\
        &=U\parens{l, l+\frac{n+n_0+1}{n+n_0}w_n}.
       \end{align*}

       Note that the updated uniform posterior predictive distribution has a wider range
       than the empirical distribution $U(l,l+w_n)$,
       thus ``has the ability to extrapolate from the data'' \cite{tenenbaum1998bayesian}.
 \item A non-informative improper prior is obtained by $n_0\downto 0$:

\begin{align*}
 p(w)\sim w^{-n_0-1}
 &\to[n_0\downto 0] w^{-1}.\\
 p(w|X)=\pareto(n_0+n, w_n)
 &\to[n_0\downto 0] \pareto(n, w_n).
\end{align*}

\end{enumerate}

\newpage

\subsection{Uniform $U(l,u=l+w)$ with Unknown Lower Bound $l$}

\begin{ex}
The Allies have captured $N=100$ latest Nazi tanks each of which has a serial number painted on the side.
We know they produced $u=10000$ tanks in total.
The minimum number on these latest tanks that we observed so far is $\el=1945$.
When did they stop producing the older version?
\end{ex}

\begin{enumerate}
 \setcounter{enumi}{1}
 \item Latents: $l$.
 \setcounter{enumi}{3}
 \item Distribution family and parameters:
       \begin{enumerate}
        \item $p(x_i|l)=U(x_i|l,l+w). \quad (w>0)$
        \item $p(l)=U(u_0-w, l_0).\quad (u_0-w<l<l_0)$.
        \item $p(l|X)=U(u_n-w, l_n)$.
       \end{enumerate}
 \setcounter{enumi}{5}
 \item $p(x_i|l)=w^{-1}$, $l<x_i<l+w$.
 \item $p(X|l)=w^{-n}$, $l<\min_i x_i < \max_i x_i<l+w$.

       Let $\eu=\max_i x_i$ and $\el=\min_i x_i$.
       Then $\eu-w<l<\el$.
 \item $p(l)=(l_0-u_0+w)^{-1}$.
 \item Using the second strategy.
       $p(l,X)=p(X|l)p(l)=\text{Const}.$ where $\max(u_0,\eu)-w<l<\min(l_0,\el)$,
       otherwise 0.
       Let $u_n=\max (u_0, \eu)$, $l_n=\min (l_0, \el)$.
       \begin{align*}
        p(X)
        &
        \textstyle=\int_\R p(l,X)dl
        \textstyle=\int_{u_n-w}^{l_n} p(l,X)dl\\
        &
        \textstyle=\brackets{w^{-n}(l_0-u_0+w)^{-1}\cdot l}_{u_n-w}^{l_n}
        \textstyle=w^{-n}\frac{l_n-u_n+w}{l_0-u_0+w}.\\
        p(l|X)
        &=\frac{p(l,X)}{p(X)}=(l_n-u_n+w)^{-1} = U(u_n-w,l_n).
       \end{align*}
 \setcounter{enumi}{4}
 \item $p(x,l|X)=p(x|l)p(l|X) = w^{-1}(l_n-u_n+w)^{-1}.$
       Note that $l<x<l+w$ for $p(x|l)$, thus $x-w<l<x$. Therefore
       \begin{align*}
        p(x|X)
        &=\int_{\max(u_n, x)-w}^{\min(l_n,x)} p(x,l|X)dl\\
        &=\frac{\min(l_n,x)-\max(u_n, x)+w}{w(l_n-u_n+w)}\\
        &\hspace{-6em}=\left\{
        \begin{array}{cl}
         \frac{l_n-x+w}{w(l_n-u_n+w)},& (l_n<x\land u_n<x \iff \max(l_n,u_n)<x)\\
         w^{-1}                      ,& (l_n>x\land u_n<x \then u_n<l_n \then \bot )\\ 
         \frac{l_n-u_n}{w(l_n-u_n+w)},& (l_n<x\land u_n>x \iff l_n<x<u_n)\\
         \frac{x-u_n+w}{w(l_n-u_n+w)}.& (l_n>x\land u_n>x \iff x<\min(l_n,u_n))
        \end{array}
        \right.
       \end{align*}
       Using $p(x|X)\geq 0$,
       \begin{align*}
        \hspace{-2em}p(x|X)
        &=\left\{
        \begin{array}{cc}
         \frac{l_n-x+w}{w(l_n-u_n+w)},& (\max(l_n,u_n)<x<l_n+w)\\
         \frac{l_n-u_n}{w(l_n-u_n+w)},& (l_n<x<u_n)\\
         \frac{x-u_n+w}{w(l_n-u_n+w)}.& (u_n-w<x<\min(l_n,u_n))
        \end{array}
        \right.
       \end{align*}
       This predictive distribution has a trapezoidal shape and,
       as usual, has a support wider than the empirical distribution $U(l_n,u_n)$.

 \item A non-informative improper prior is obtained by $l_0\to-\infty, u_0\to\infty$.

\end{enumerate}

\newpage

\subsection{Uniform $U(l,u=l+w)$ with Unknown $l,w$}

\begin{enumerate}
 \setcounter{enumi}{1}
 \item Latents: $l,w$.
 \setcounter{enumi}{3}
 \item Distribution family and parameters:
       \begin{enumerate}
        \item $p(x_i|l,w)=U(x_i|l,l+w). \quad (w>0)$
        \item $p(w)=\pareto(w|n_0,w_0), \quad (w_0<w)$

              $p(l|w)=U(l|u_0-w, l_0).\quad (u_0-w<l<l_0)$

       \end{enumerate}
 \setcounter{enumi}{5}
 \item $p(x_i|l,w)=w^{-1}$, $l<x_i<l+w$.
 \item $p(X|l,w)=w^{-n}$, $l<\min_i x_i < \max_i x_i<l+w$.

       Let $\eu=\max_i x_i$, $\el=\min_i x_i$, $\ew=\eu-\el$.

       Then $\eu-w<l<\el$ and $\ew<w$.
 \item $p(l,w)=p(l|w)p(w)=(l_0-u_0+w)^{-1}\cdot n_0w_0^{n_0} w^{-n_0-1}$

       $=(l_0-u_0+w)^{-1}\cdot A w^{-n_0-1}. \quad (A=n_0w_0^{n_0})$
 \item Using the second strategy.

       Let $u_n=\max (u_0, \eu)$, $l_n=\min (l_0, \el)$, $w_n=u_n-l_n$.
       (This implies $w_0=u_0-l_0$, but this is rather accidental.)
       \begin{align*}
        p(l,w,X)&=p(X|l,w)p(l|w)p(w)\\
        &=w^{-n}\cdot (w-w_0)^{-1} \cdot A w^{-n_0-1}\\
        &=A w^{-(n+n_0+1)}(w-w_0)^{-1}.\\
        p(w,X)
        &=\int_{u_n-w}^{l_n} p(l,w,X)dl
         =A w^{-(n+n_0+1)}\frac{w-w_n}{w-w_0}.\\
        p(l|w,X)
        &=\frac{p(l,w,X)}{p(w,X)}=\parens{w-w_n}^{-1}
         =U(l|u_n-w,l_n).
       \end{align*}

       Let $z=w^{-1}$, $z_n=w_n^{-1}$, $w\in[w_n,\infty]$, $z\in [0,z_n]$, $N=n+n_0$.
       Then $p(X)=\int_{w_n}^\infty p(w,X)dw$
       \begin{align*}
        &\hspace{-2em}\textstyle=\int_{z_n}^0 A z^{N+1}\frac{z^{-1}-z_n^{-1}}{z^{-1}-z_0^{-1}}\parens{\frac{dz}{-z^2}}
         \textstyle=\int_{z_n}^0-A z^{N-1}\frac{z-z_n}{z-z_0}\frac{-zz_0}{-zz_n}dz\\
        &\hspace{-2em}\textstyle=\int_0^{z_n} \parens{\frac{z^{N}}{z-z_0}-\frac{z_nz^{N-1}}{z-z_0}} A\frac{z_0}{z_n} dz
         \textstyle=\int_0^{\frac{z_n}{z_0}} \parens{\frac{z_0}{z_n} \frac{t^{N}}{1-t} - \frac{t^{N-1}}{1-t}} A\frac{z_0^{N}}{z_n} dt\\
        &\hspace{-2em}\textstyle=\parens{\frac{z_0}{z_n} \Beta\parens{\frac{z_n}{z_0}; N+1, 0} - \Beta\parens{\frac{z_n}{z_0}; N, 0}}A\frac{z_0^{N}}{z_n}\\
        &\hspace{-2em}\textstyle=\parens{\frac{w_n}{w_0} \Beta\parens{\frac{w_0}{w_n}; N+1, 0} - \Beta\parens{\frac{w_0}{w_n}; N, 0}}A\frac{w_n}{w_0^{N}}
         \textstyle=AC(N).\\
        &\textstyle p(w|X)=\frac{p(w,X)}{p(X)}=w^{-(N+1)}\frac{w-w_n}{w-w_0} C(N)^{-1}.
       \end{align*}
       Note that the posterior $p(w|X)$ is not conjugate with $p(w)=\pareto(w|n_0,w_0)$.
       However, as we see below, this does not affect the posterior predictive $p(x|X)$.

 \setcounter{enumi}{4}
 \item $p(x|X)=\iint p(x|w,l)p(l|w,X)p(w|X)dldw$
       \begin{align*}
        &\hspace{-2em}\textstyle =\iint w^{-1} \cdot (w-w_n)^{-1} \cdot w^{-(N+1)}\frac{w-w_n}{w-w_0} C(N)^{-1} dldw\\
        &\hspace{-2em}\textstyle =\iint \frac{w^{-(N+2)}}{w-w_0} C(N)^{-1} dldw
         \textstyle =\int w^{-(N+2)}\frac{w-w_n}{w-w_0} C(N)^{-1} dw \\
        &\hspace{-2em}\textstyle = \frac{C(N+1)}{C(N)} = U(x|u_n-\frac{C(N+1)}{C(N)},u_n).\\
        &\textstyle\frac{C(N+1)}{C(N)}=w_0^{-1}\frac{\frac{w_n}{w_0} \Beta\parens{\frac{w_0}{w_n}; N+2, 0} - \Beta\parens{\frac{w_0}{w_n}; N+1, 0}}{\frac{w_n}{w_0} \Beta\parens{\frac{w_0}{w_n}; N+1, 0} - \Beta\parens{\frac{w_0}{w_n}; N, 0}}.
       \end{align*}

\end{enumerate}

\newpage

\end{document}